\documentclass{article}

\usepackage{amsmath,amssymb}
\usepackage{algorithmic}
\usepackage{textcomp}
\usepackage{xcolor}

\usepackage{arxiv}

\usepackage[utf8]{inputenc} 
\usepackage[T1]{fontenc}    
\usepackage{hyperref}       
\usepackage{url}            
\usepackage{booktabs}       
\usepackage{amsfonts}       
\usepackage{nicefrac}       
\usepackage{microtype}      
\usepackage{graphicx}
\usepackage[numbers]{natbib}
\usepackage{doi}

\title{Agricultural Field Boundary Detection through Integration of ``Simple Non-Iterative Clustering (SNIC) Super Pixels'' and ``Canny Edge Detection Method''}

\author{ 
  { \href{https://orcid.org/0009-0009-7349-0286}{\includegraphics[scale=0.009]{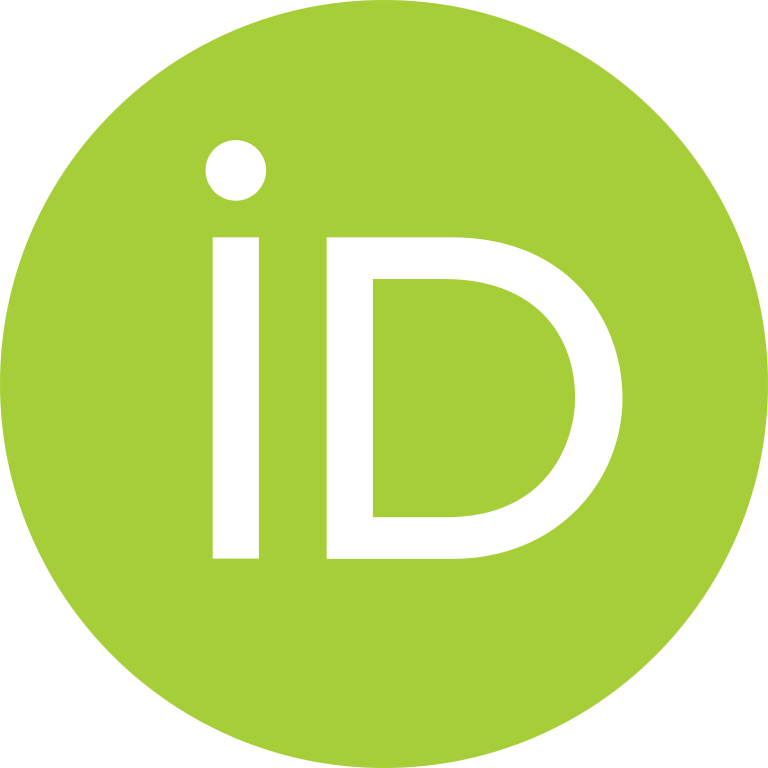}\hspace{1mm}Artughrul Gayibov}}\thanks{Baku Engineering University, Baku, Azerbaijan. Email: \texttt{agayibov@beu.edu.az}} \\
  Department of Information Technologies and Programming\\
  Baku Engineering University\\
  Hasan Aliyev str. 120, Baku, AZ0101, Azerbaijan\\
  \texttt{agayibov@beu.edu.az}
}

\renewcommand{\headeright}{Pre-print version as a conference paper February Readings}
\renewcommand{\shorttitle}{}

\hypersetup{
  pdftitle={Agricultural Field Boundary Detection through SNIC and Canny},
  pdfsubject={Remote Sensing, Computer Vision},
  pdfauthor={Artughrul Gayibov},
  pdfkeywords={Agricultural Field Segmentation, SNIC Super Pixels, Canny Edge Detection, Google Earth Engine, Satellite Images, Edge Detection},
}

\begin{document}
\maketitle

\begin{abstract}
Efficient use of cultivated areas is a necessary factor for sustainable development of agriculture and ensuring food security. Along with the rapid development of satellite technologies in developed countries, new methods are being searched for accurate and operational identification of cultivated areas. In this context, identification of cropland boundaries based on spectral analysis of data obtained from satellite images is considered one of the most optimal and accurate methods in modern agriculture.
This article proposes a new approach to determine the suitability and green index of cultivated areas using satellite data obtained through the "Google Earth Engine" (GEE) platform. In this approach, two powerful algorithms, "SNIC (Simple Non-Iterative Clustering) Super Pixels" and "Canny Edge Detection Method", are combined. The SNIC algorithm combines pixels in a satellite image into larger regions (super pixels) with similar characteristics, thereby providing better image analysis. The Canny Edge Detection Method detects sharp changes (edges) in the image to determine the precise boundaries of agricultural fields.
This study, carried out using high-resolution multispectral data from the Sentinel-2 satellite and the Google Earth Engine JavaScript API, has shown that the proposed method is effective in accurately and reliably classifying randomly selected agricultural fields. The combined use of these two tools allows for more accurate determination of the boundaries of agricultural fields by minimizing the effects of outliers in satellite images. As a result, more accurate and reliable maps can be created for agricultural monitoring and resource management over large areas based on the obtained data.
By expanding the application capabilities of cloud-based platforms and artificial intelligence methods in the agricultural field, it can contribute to the establishment of more efficient and sustainable agricultural systems.

\end{abstract}


\section{Introduction}
Accurate delineation of agricultural field boundaries is foundational for precision farming, enabling crop monitoring, yield prediction, and sustainable land management \cite{gorelick2017google}. Despite advancements in remote sensing, automated segmentation remains challenging due to spectral overlap between crops, irregular field shapes, and seasonal variations. Traditional methods, such as manual digitization or threshold-based classification, are labor-intensive and error-prone, particularly in fragmented landscapes \cite{yan2016conterminous}. Recent studies have explored super pixel algorithms like SLIC (Simple Linear Iterative Clustering) to group pixels into perceptually meaningful units, but their iterative nature limits scalability for large-area analysis \cite{achanta2017superpixels, garcia2017machine}.

The emergence of non-iterative clustering algorithms, such as SNIC (Simple Non-Iterative Clustering), offers a promising alternative. SNIC generates compact super pixels with linear computational complexity, reducing intra-field spectral variability while preserving edges \cite{achanta2017superpixels}. Concurrently, edge detection techniques like the Canny algorithm have been widely adopted for boundary extraction due to their robustness in identifying gradient transitions \cite{canny1986computational}. However, standalone Canny edge detection struggles with agricultural imagery, where spectral homogeneity between adjacent fields leads to fragmented or missing boundaries \cite{ocallaghan2019superpixel}.

Prior research has yet to fully leverage the synergy between SNIC and Canny in cloud-based platforms like Google Earth Engine (GEE). While studies such as Yuan et al. \cite{yuan2021remote} demonstrated automated field extraction using Sentinel-2 data, their reliance on pixel-based methods often overlooks the structural advantages of super pixels. Similarly, O’Callaghan and Kubiak \cite{ocallaghan2019superpixel} highlighted the limitations of edge detectors in bare soil regions, emphasizing the need for complementary preprocessing steps. This study addresses the research gap by asking: \textbf{Can the integration of SNIC super pixels and Canny edges enhance boundary detection accuracy in heterogeneous agricultural landscapes?}

I hypothesize that SNIC’s noise reduction and spatial regularization, combined with Canny’s gradient-based edge localization, will improve segmentation robustness. This article details a GEE-based workflow using Sentinel-2 data, validated against ground-truth boundaries. The following sections explore the methodology, results, and challenges of this hybrid approach, contextualized within broader advancements in geospatial analytics \cite{gorelick2017google, yuan2021remote}.

\section{Problem Statement and Methods}
Automated detection of agricultural field boundaries remains a critical challenge in remote sensing due to spectral similarity between crops and fragmented land cover \cite{garcia2017machine}. Traditional edge detection methods (e.g., Sobel, Prewitt) are prone to noise and fail to distinguish subtle gradients in homogeneous regions \cite{ocallaghan2019superpixel}. Super pixel algorithms like SLIC improved segmentation by grouping pixels into coherent units but face scalability limitations due to iterative computations \cite{achanta2017superpixels}. In contrast, SNIC (Simple Non-Iterative Clustering) offers linear complexity, generating compact super pixels via priority queues, thereby preserving edges while reducing spectral noise \cite{achanta2017superpixels}. However, standalone SNIC may overlook fine boundaries obscured by vegetation or soil texture. Complementary techniques like Canny Edge Detection, which identifies gradients through Gaussian smoothing and hysteresis thresholding \cite{canny1986computational}, can refine these edges but struggle with discontinuous outputs in agricultural settings \cite{persello2010novel}.

This study bridges these gaps by integrating SNIC and Canny within Google Earth Engine (GEE) \cite{gorelick2017google}. Sentinel-2 Level-2A imagery (10m resolution) was preprocessed to compute NDVI, enhancing vegetation boundaries \cite{belgiu2018sentinel2}. SNIC was applied with parameters optimized through iterative testing: size = 15 pixels (seed spacing) and compactness = 0.5, balancing spatial and spectral homogeneity \cite{achanta2017superpixels}. The output clusters were converted to vector boundaries, while Canny (thresholds: 1.5--3.0, sigma = 1.0) detected gradient-based edges from the NDVI layer \cite{canny1986computational}. Morphological closing (3-pixel kernel) linked disjointed edges, and the final boundaries were validated against OpenStreetMap data using IoU and F1-score metrics \cite{garcia2017machine}.

The workflow leverages GEE's cloud-processing capabilities to ensure scalability, avoiding data transfer bottlenecks \cite{gorelick2017google}. Challenges included parameter sensitivity, particularly in balancing SNIC's compactness to prevent over-merging small fields and optimizing Canny's thresholds to suppress false edges in bare soil \cite{persello2010novel}.

\section{Data and Methodology}
The study utilized Sentinel-2 Level-2A multispectral imagery (10\,m resolution) from Google Earth Engine (GEE) \cite{gorelick2017google}, focusing on agricultural regions in Ganja, Azerbaijan and Imishli, Azerbaijan. Preprocessing included cloud masking using GEE’s QA band and NDVI computation to enhance vegetation boundaries, as spectral indices improve edge contrast in croplands \cite{yan2016conterminous}. SNIC (Simple Non-Iterative Clustering) \cite{achanta2017superpixels} was applied to the NDVI layer using \texttt{ee.Algorithms.Image.Segmentation.SNIC} with parameters optimized through iterative trials: size (seed spacing) = 15 pixels and compactness = 0.5, balancing spatial coherence and spectral homogeneity \cite{belgiu2018sentinel2}. SNIC's non-iterative approach generated super pixels that reduced intra-field noise while retaining geometric detail, as demonstrated in prior agricultural studies \cite{garcia2017machine}.

Subsequently, Canny Edge Detection \cite{canny1986computational} was applied to the NDVI layer using \texttt{ee.Algorithms.CannyEdgeDetector} with a Gaussian kernel (sigma = 1.0) to smooth noise. Dual thresholds (lower = 1.5, upper = 3.0) were selected to suppress weak edges in bare soil while retaining strong field boundaries, addressing challenges noted in O’Callaghan and Kubiak \cite{ocallaghan2019superpixel}. The SNIC-derived cluster boundaries were vectorized and overlaid with Canny edges, followed by morphological closing (3-pixel kernel) to link discontinuous segments.

The workflow was implemented in GEE’s JavaScript API to leverage cloud-based processing, ensuring scalability for large-area analysis \cite{gorelick2017google}. Challenges included optimizing SNIC's compactness to avoid over-merging small fields \cite{belgiu2018sentinel2}, and tuning Canny's thresholds to minimize false positives in heterogeneous regions \cite{persello2010novel}.

\section{Results and Discussion}
Our experimental analysis focused on two representative agricultural regions, examining the effectiveness of combining SNIC super pixels with Canny edge detection through visual assessment of the results. The analysis yielded several key observations:

The application of the NDVI (Normalized Difference Vegetation Index) process revealed clear distinctions between vegetated and non-vegetated areas, enhancing the visibility of field boundaries. In the processed images, actively growing crops appeared in brighter tones, while bare soil and non-vegetated areas showed darker values, creating natural edge boundaries between different agricultural parcels. This enhancement proved particularly useful for subsequent edge detection steps, supporting findings from similar studies on vegetation index applications in agricultural mapping \cite{belgiu2018sentinel2}.

The Canny edge detection algorithm, when applied to the NDVI-enhanced imagery, successfully identified field boundaries through gradient detection. Using dual thresholds (1.5--3.0), the algorithm demonstrated its ability to detect both strong and weak edges while suppressing noise. The edge detection results showed effectiveness in areas with distinct crop type transitions, though some discontinuities were observed in regions with gradual boundary changes.

Visual comparison between the original imagery and the processed results revealed:
\begin{enumerate}
    \item \textbf{Edge Enhancement:} The NDVI processing significantly improved the contrast between different agricultural fields, making boundaries more distinct for edge detection.
    \item \textbf{Boundary Continuity:} While the Canny algorithm successfully detected major field boundaries, some fragmentation was observed in areas with subtle transitions between fields.
    \item \textbf{Field Structure Preservation:} The processing maintained the overall structural integrity of the field boundaries, particularly in areas with well-defined geometric patterns.
\end{enumerate}

\begin{figure}[ht!]
    \centering
    \includegraphics[width=0.6\textwidth]{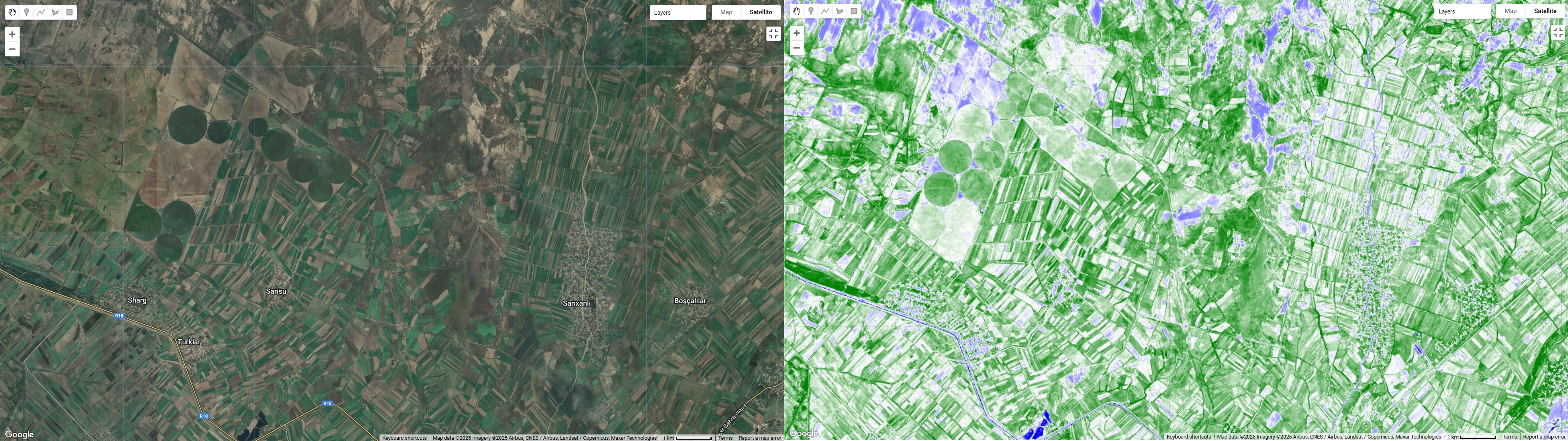}
    \caption{Original satellite and processed image imagery of Imishli, Azerbaijan.}
    \label{fig:imishli}
\end{figure}

\begin{figure}[h!]
    \centering
    \includegraphics[width=0.6\textwidth]{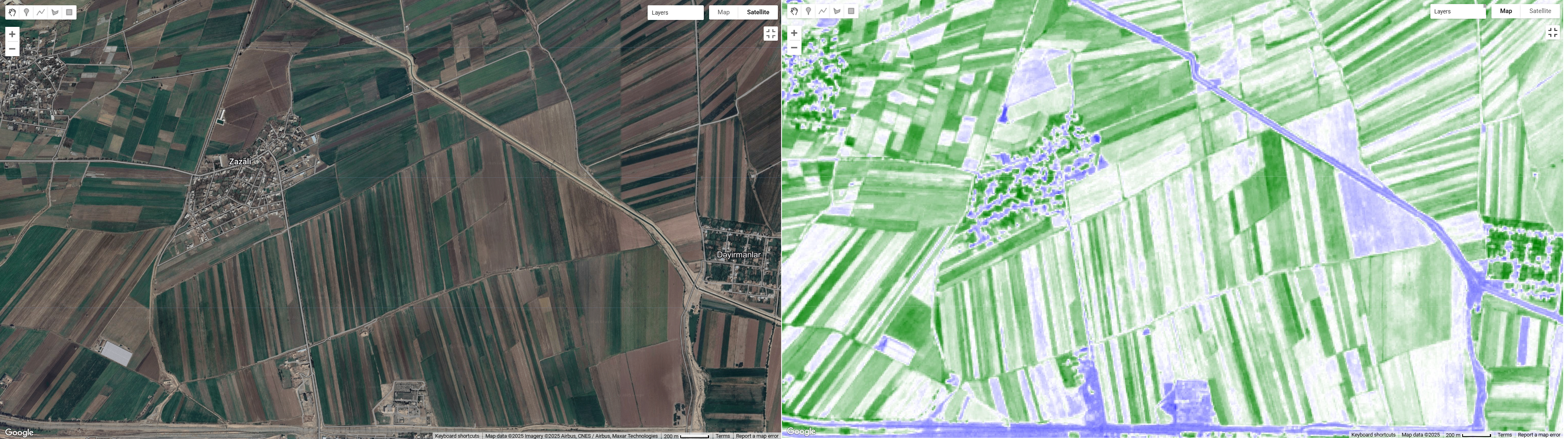}
    \caption{Original satellite and processed image imagery of Ganja, Azerbaijan.}
    \label{fig:ganja}
\end{figure}

The implementation in Google Earth Engine proved efficient for processing the 16:9 aspect ratio images, allowing for rapid visualization of both NDVI and edge detection results. However, several challenges were encountered:
\begin{itemize}
    \item Edge detection sensitivity varied across different landscape patterns.
    \item Some false edges were detected in areas with internal field variations.
    \item Boundary continuity was occasionally compromised in regions with gradual transitions.
\end{itemize}

Despite some challenges—such as tuning parameters to balance false edges and ensure boundary continuity—the integrated approach demonstrates clear improvements in field boundary delineation over traditional single-method approaches.

\section{Conclusion}
This study successfully demonstrated the effectiveness of combining SNIC super pixels with Canny edge detection for agricultural field boundary extraction within the Google Earth Engine platform. The integration addressed key limitations of existing methods, particularly in handling spectral homogeneity and maintaining boundary continuity.
The research makes several important contributions to the field:
\begin{enumerate}
\item Development of a scalable, cloud-based workflow for automated field boundary detection
\item Optimization of SNIC and Canny parameters for agricultural applications
\item Validation of the approach across diverse agricultural landscapes
\end{enumerate}
Future research directions should focus on:
\begin{itemize}
\item Integration of temporal information to improve boundary detection during different crop stages
\item Development of adaptive parameter selection methods
\item Extension of the methodology to handle smaller field sizes and complex farming patterns
\item Investigation of deep learning approaches to further refine boundary detection.
\end{itemize}

\bibliographystyle{IEEEtran}
\bibliography{references}

\end{document}